\documentclass[reqno,11pt]{article}

\usepackage{authblk}
\usepackage{color}
\usepackage{hyperref}
\usepackage{amsmath}
\usepackage{amsfonts}
\usepackage{amssymb}
\usepackage{graphicx}
\usepackage{amsthm}
\usepackage{natbib}
\usepackage{url}
\usepackage {tikz}
\usetikzlibrary {positioning}
\usetikzlibrary{decorations.shapes}
\definecolor {processblue}{cmyk}{0.96,0,0,0}

\title{Learning to Ask Medical Questions using Reinforcement Learning}
\author[1]{Uri Shaham}
\author[2]{Tom Zahavy}
\author[1]{Cesar Caraballo}
\author[1]{Shiwani Mahajan}
\author[1]{Daisy Massey}
\author[1]{Harlan Krumholz}
\affil[1]{Center for Outcomes Research and Evaluation, Yale University}
\affil[2]{Department of Electrical Engineering, Technion}

\date{}                                           

\begin{document}
\maketitle

\begin{abstract} 
We propose a novel  reinforcement learning-based approach for adaptive and iterative feature selection.
Given a masked vector of input features, a reinforcement learning agent iteratively selects certain features to be unmasked, and uses them to predict an outcome when it is sufficiently confident. 
The algorithm makes use of a novel environment setting, corresponding to a non-stationary Markov Decision Process. A key component of our approach is a guesser network, trained to predict the outcome from the selected features and parametrizing the reward function.
Applying  our method to a national survey dataset, we show that it not only outperforms strong baselines when requiring the prediction to  be made based on a small number of input features, but is also highly more interpretable. Our code is publicly available at \url{https://github.com/ushaham/adaptiveFS}.

\end{abstract} 

\section {Introduction}\label{sec:intro}
Feature selection is an important topic in traditional machine learning~\citep{li2018feature}, which motivated a large number of widely adopted works, e.g., Lasso~\citep{tibshirani1996regression}.
In various cases, the process of obtaining input measurements requires considerable effort (e.g., time, money, technology). 
For example, in medical datasets input features may correspond to lab tests, medical imaging results, or even answers to questionnaires, which are expensive and slow to produce.
Allowing oneself to be able to accurately predict a response variable from a small set of input features is thus a desirable goal, which can be manifested in saving time, money, and sometimes even human lives.

As a running example, consider the case of a patient complaining to a family doctor about not feeling well.
The doctor then asks the patient several questions about his current condition and medical background, and may also ask the patient to do some lab tests. 
Implicitly, the doctor is aiming at quickly collecting relevant details on the patient that will allow her to have a clear understanding of the patients' medical status, and consequently decide on an appropriate action (e.g., medication prescription, admit to hospitalization etc.).
The doctor would keep asking questions  as long as this improves her understanding of the patient's medical status. 
Whenever the information acquired so far enables her to obtain a clear picture of the patient's status, she can make a decision about the next required steps. 
Making a (good) decision early in the process is beneficial, for example if the patient needs urgent treatment or when it is resourceful to acquire additional information (e.g., lab tests, medical diagnostics).  
Hence, a first challenge would be to navigate through the trade-off of gathering sufficient information while doing so as quickly as possible. 
Considering this example, it is also straightforward to realize that it would be highly sub-optimal to always select the same small subset of input features, regardless of the patient and the complaint. 
Indeed, when a 83 year old male complains about headache, we would expect the doctor to choose a different investigation path, comparing to a case of 7 year old girl complaining on stomachache.  
Put differently, it is often desirable to select the input features adaptively.
Moreover, in cases like this it is also natural to select features iteratively, so that the $k$'th feature is selected after the values of the previous $k-1$ selected ones are known.

In this manuscript we aim at these desired attributes and propose a novel reinforcement learning (RL)-based approach for adaptive and iterative feature selection.
In our RL framework, the state corresponds to the current agent's perception of the input sample, and the action space corresponds to the set of available input features. At each time step through an episode, a RL agent chooses a feature whose value is masked or unknown, and gets to observe the value of the specific feature. Once confident, the agent may choose to predict the label of the input sample and is rewarded based on the quality of the prediction. 
Throughout training, the agent learns to ``ask'' for the features which are most helpful for an accurate prediction of the label.
The set of selected features may differ for each input sample, and the features are selected gradually and in adaptive fashion, so that each feature is selected based on the previously selected ones and their corresponding values.
In this sense, the agent behaves in a more human-like fashion, comparing to standard feature selection models.
Moreover, the trajectory (of selected features and their corresponding values) leading to each prediction is fully transparent and hence contributes to the model's interpretability.

We apply our approach to a national survey dataset, containing answers of patients to a large set of questions.  We demonstrate that when limiting the prediction to be based on a small number of features, our approach outperforms strong off-the-shelf baselines, while also being more interpretable.

Our contributions are threefold:
From a technical perspective, we apply reinforcement learning for adaptive feature selection, and propose a novel environment design to support it. 
Doing so, we propose a non-stationary Markov Decision Process (MDP) setting, in which the reward function is learned. 
From a medical perspective, we show how to design a human-like AI interface which adaptively selects information pieces in order to predict 4-year mortality. 
From an experimental perspective, we show that our approach outperforms strong off-the-shelf baselines, while also being more interpretable.

The remainder of this manuscript is organized as follows: 
In Section~\ref{sec:related} we review related literature.
 The data cohort is described in Section~\ref{sec:cohort}. 
Our proposed approach is presented in Section~\ref{sec:afs}. 
In Section~\ref{sec:experimental} we report experimental results. 
Section~\ref{sec:conclusions} briefly concludes the manuscript.

\section {Related Work}\label{sec:related}

As the main motivation behind choosing a RL approach for the current adaptive feature selection task serve several recent applications of RL to the 20 questions game.
In this game the player's goal is to predict the identity of an unknown character, where in each step of the game the player chooses a Yes/No question to ask and obtains the corresponding answer.
  
\citep{hu2018playing} define their action space as the set of possible questions and the state as a distribution over the possible characters. Their state update mechanism uses statistics of people's responses to questions.
In addition, as a non-zero reward is obtained only at the end of the episode (corresponding to game win / lose), they augment their environment with a reward network, generating  an intrinsic immediate reward at every time step, whose goal is to predict the true reward. 

\citep{chen2018learning} use an Long Short Term Memory (LSTM) state update mechanism, so that the state space is the hidden space of the LSTM network. They use a naive-Bayes mechanism for making predictions at the end of each episode. 
In addition, their approach consists of two major elements: a DQN RL agent who plays the game and a knowledge acquisition mechanism, whose goal is to estimate answers distributions to questions (regardless of the specific episode being played), which utilizes a matrix decomposition mechanism.  

\citep{zhao2016towards} propose a RL-based dialogue state tracking system, which they apply to the 20 questions game. Their approach combines reward signal with label supervision, which is related to our approach, as will be explained in Section~\ref{sec:proposed}. 

Several works have focused on integrating feature selection methods with deep learning, e.g., ~\citep{li2016deep, roy2015feature, zhao2015heterogeneous, louizos2017learning, yamada2018deep, balin2019concrete}.
However, these approaches are neither adaptive nor iterative, which are core requirements in the setting we consider in this manuscript.

Lastly, the topic of intrinsic reward design (see, for example~\cite{zheng2018learning}), used also by~\citep{hu2018playing}, has drawn much interest in the RL community. It involves design of intrinsic reward functions, which guiding the agent towards maximizing expected external reward (which may be sparse, or obtained at late time steps, for example). 
This has relations with our proposed approach, where we train and use use a guesser network to provide rewards to the RL agent, as will be explained in Section~\ref{sec:proposed}.

\section {Cohort}\label{sec:cohort}
In this section we describe the data cohort we apply our method to. Detailed instructions for data acquisition and preprocessing can be found in Appendix~\ref{sec:dpw}.

\subsection {Data Source}\label{sec:dsource}
We used 10-year data (2002 to 2011) from the National Health Interview Survey (NHIS), which is an annual cross-sectional survey conducted by the National Center for Health Statistics (NCHS) that provides estimates on the health status, health-care access, and behaviors of the non-institutionalized US population\footnote{\url{https://www.cdc.gov/nchs/nhis/about_nhis.htm}}.
The sample design of the NHIS follows a multistage area probability design, adjusting for non-response and allowing for nationally representative sampling of households and individuals. This sample design includes under-represented groups. The NHIS questionnaire is divided into 4 cores: Household Composition core, Family core, Sample Child core, and Sample Adult core. The Household Composition core includes basic and relationship information about all individuals in the household. The Family Core file includes socio-demographic characteristics, health insurance coverage, basic indicators of health status, injuries, activity limitations, and access to and utilization of health care services separately for each family in the household. A random child and adult from each family are selected to gather more detailed information about them, constituting the Sample Child and the Sample Adult cores, respectively. In our study, we used the Sample Adult core files with variables supplemented from the other cores. The Household response rates ranged from 89.6\% in 2002 to 79.5\% in 2011. All survey participants provided informed consent before participation in the survey. This study received exemption from Yale University Institutional Review Board Committee because NHIS data are publicly available and de-identified.

The NHIS Linked Mortality files include data from all surveys between 1985 and 2014, linked to the National Death Index (NDI), with follow-up to the date of death or 31 December 2015\footnote{\url{https://www.cdc.gov/nchs/data-linkage/mortality-public.htm}}.
An estimated 95.4\% of baseline participants had the eligible mortality follow-up information. The NCHS at the Centers for Disease Control and Prevention used post-stratification re-weighting based on the U.S. population to account for ineligible follow-up\footnote{\url{https://www.cdc.gov/nchs/data-linkage/mortality-methods.htm}}. The NDI file provides data on the mortality status, year of death, quarter of year, and cause of death (categorized into the following groups based on ICD-9 and ICD-10 codes – heart disease, cancer, chronic lower respiratory disease, cerebrovascular diseases, diabetes, pneumonia and influenza, Alzheimer’s disease, kidney disease, and unintentional injuries).

\subsection {Study Population}\label{sec:spop}
A total of 282,001 adults aged 18 years and above were interviewed between 2002-2011, of which we excluded those with missing information on their mortality follow-up ($n=12,905$) resulting in a study sample of $n=269,069$ individuals

\subsection {Outcome Definition}\label{sec:outcome}
Our outcome of interest was death within 4 years of the date of interview. We used 16 quarters of a year from the quarter of the interview to the quarter of death as our follow-up time, as the NHIS Linked Mortality files provides the year and quarter of death only. 

\subsection {Candidate Variables}\label{sec:candid}
Because some of the content of the NHIS questionnaire changed over the years depending on the data needs for current health topics, we decided to include questions that were consistent across all 10 years (1,022 variables out of 2,360 total). We then excluded those variables that were only asked to the participant conditional on a prior answer to another question (811 variables, labeled as “daughter variables”), those that had $>80$\% missingness (8 variables), and those that contained redundant information with other variables (for example, identifiers, repeated information; 45 variables). The final dataset had 211 variables, of which 188 were interview variables (candidate variables for the model) and 23 were identifiers and outcome variables. For the 10 continuous variables with missing values (ranging from 1\%–27\% missing rates), we used single-value imputation with median. For 11 categorical variables with missing values (ranging from 3\%–78\% missing rates) we created “missing” as a separate category. Finally, we added the daughter questions (85 variables) for the top 25 variables most correlated with outcome and for the top 25 variables identified using an XGBoost model, increasing the total number of candidate variables to 273 variables. Categorical variables were one-hot encoded.

\section {Adaptive Feature Selection using Reinforcement Learning}\label{sec:afs}

\subsection {Preliminaries}\label{sec:preliminaries}
\subsubsection {Reinforcement Learning}\label{sec:rl}

Reinforcement Learning (RL) is family of algorithms aimed at solving Markov Decision Processes (MDPs).
A MDP is represented as a tuple $(\mathcal{S}, \mathcal{A}, \mathcal{P}, \mathcal{R}, \gamma)$,
where $\mathcal{S}$ is a set of states (also called state space), $ \mathcal{A}$ is a set of actions (also called action space), $\mathcal{P}$ is a set of state transition rules
$$\mathcal{P}(s', s, a) = \mbox{Prob}(S_{t+1} = s' | S_t = s, A_t = a),$$
specifying the distribution over the state space for the next state given a current state and action, $\mathcal{R}:\mathcal{S}\times \mathcal{A} \rightarrow \mathbb{R}$ is a reward function and $0 < \gamma \le 1$ is a discount factor.

The RL paradigm consists of two major elements: an agent and an
environment. 
The agent follows a policy, defined as a function $ \pi:\mathcal{S} \rightarrow \mbox{dist}(\mathcal{A})$ 
which maps each state to a distribution over the action space.
Doing so, it interacts with the environment by choosing actions from $\mathcal{A}$ that may let it move between states. 
At each step $t$ of the episode, being in state $s_t$, the agent chooses an action $a_t$ from $\pi(s_t)$, obtains a (negative, zero or positive) reward $r_t=\mathcal{R}(s_t, a_t)$ and moves to state $s_{t+1}$, which is sampled from  $\mathcal{P}(\cdot, s_t, a_t)$.

The agent's goal is to learn a policy that maximizes the expected reward
$$ \arg\max_\pi \mathbb{E}[R_T] = \arg\max_\pi \mathbb{E}\left[\sum_{t=1}^T\gamma^t r_t  \biggm\lvert a_t \sim \pi(s_t), \, s_{t+1} \sim\mathcal{P}(\cdot, s_t, a_t)\right] ,$$
where $T$ is the length of the episode.

Rather than supervised learning, in which ground truth labels
are known during training, such knowledge is absent in RL. 
Instead, the reward signal is the driving force of the learning. 
This weaker form of supervision signal makes RL systems take longer to train comparing to supervised learning algorithms. However, it is applicable to many scenarios where supervised learning is not an available approach.
In recent years, RL has been an active research area in the machine learning community, and some dramatic successes demonstrated its potential, e.g.,~\citep{silver2016mastering}.

\subsubsection {$Q$ learning and Deep $Q$ Learning}\label{sec:dqn}
In reinforcement learning, the $Q$ value function $Q^\pi:\mathcal{S}\times \mathcal{A} \rightarrow \mathbb{R}$ of a policy $\pi$  corresponds to the expected reward given the current state and the current chosen action, where the agent follows $\pi$ after taking the action.
By definition, the $Q$ function satisfies a recursive relation, known as Bellman equation\citep{sutton2018reinforcement}:
\begin{equation}
Q^\pi(s, a) =\mathcal{R}(s, a) + \mathbb{E}_{s' \sim\mathcal{P}(\cdot, s, a),\, a' \sim \pi(s')} \left [\gamma Q^\pi(s', a')\right].\notag
\end{equation}
A policy $\pi^*$ maximizing $Q^\pi$ for every $s\in \mathcal{S},\, a\in\mathcal{A}$ yields the optimal $Q$ function, denoted $Q^*$, whose corresponding Bellman equation is
\begin{equation}
Q^*(s, a) =\mathcal{R}(s, a) + \mathbb{E}_{s' \sim\mathcal{P}(\cdot, s, a)} \left [\gamma \max_{a'} Q^*(s', a')\right]. \label{eq:bellman}
\end{equation}
In words, equation~\eqref{eq:bellman} means that the current expected reward equals the sum of the immediate reward and the best (over choice of action) possible expected reward of the next state. 

In $Q$ learning~\citep{watkins1992q}, the $Q$ function is iteratively updated, in order to have the Bellman equation satisfied:
\begin{equation}
Q(s, a) \leftarrow Q(s, a)  + \alpha \left[\mathcal{R}(s, a) + \gamma \max_{a'} Q(s', a') - Q(s, a)  \right].\notag
\end{equation}

Deep $Q$ network (DQN, \citep{mnih2015human}) is arguably the first major milestone in utilizing deep neural networks for reinforcement learning. It is an instance of Fitted $Q$ Iteration~\citep{ernst2005tree}, where the $Q$ function is represented by a neural network (DQN - deep $Q$ network), parametrized by $\theta$.
The net is trained to minimize the squared difference between the left and right hand sides of the Bellman Equation 
\begin{equation}
L_{DQN}(\theta) = \left( \left(\mathcal{R}(s, a) + \gamma \max_{a'} Q(s', a'; \hat{\theta})\right) - Q(s, a; \theta)\right)^2, \label{eq:loss}
\end{equation}
where $Q(\cdot, \cdot; \hat{\theta})$ is a target network, having identical architecture as the $Q$ network, and whose parameter vector $\hat{\theta}$ is copied from the $Q$ network parameter $\theta$ every certain number of training iterations.
The method makes use of Experience Replay, which is a storage buffer holding historical instances of the form $(s_t, a_t, r_t, s_{t+1})$ which are experienced by the agent during the episode. At the end of each training episode, a minibatch of such instances is randomly sampled from the buffer and uses for gradient computation, following~\eqref{eq:loss}.

Double DQN (DDQN~\citep{van2016deep}) is an improvement of DQN, aiming to be less prone to overestimation of $Q$ values. It differs from DQN in that the evaluation of the Q values and the selection of the best action uses different parameter vectors: the selection is done using the online parameter $\theta$, while the evaluation uses the target parameter vector $\hat{\theta}$.
\begin{equation}
L_{DDQN}(\theta) = \left( \left(\mathcal{R}(s, a) + \gamma  Q(s', \arg\max_a Q(s', a; \theta); \hat{\theta})\right) - Q(s, a; \theta)\right)^2. \label{eq:ddqn}
\end{equation}

\subsection {The Proposed Approach}\label{sec:proposed}

In this section we describe our proposed approach for adaptive selection of small number of questions that will allow to accurately predict the outcome variable.

\subsubsection {Rationale}\label{sec:rationale}
\begin{figure}[t]

  \centering
    \includegraphics[width=1.0\textwidth]{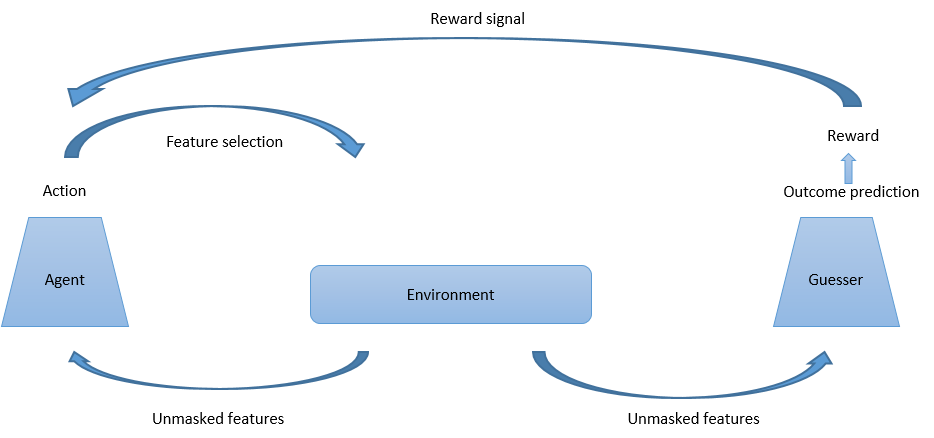}  
    \caption{The proposed approach. The agent selects features to unmask. The guesser uses the unmasked features to predict the outcome and determines the agent's reward. The agent learns to select features that will allow accurate prediction.}
    \label{fig:1}
\end{figure}

Our questionnaire dataset $D$ is a $n \times d$ matrix, where $D_{i,j}$ corresponds to the answer of patient $i$ to question $j$. 
Each episode is performed on a single patient. 
Let $x$ correspond to the ($d$-dimensional) feature vector of the patient and let $y \in \{0,1\}$ be the corresponding label.
A key component in our design is a guesser function $G: \mathcal{S} \rightarrow [0,1]$  whose goal is to predict the outcome $y$ from any state $s$, which corresponds to the unmasked entries of $x$.
At the beginning of each episode, the patient's feature vector $x$ is completely masked. 
In each time step throughout the episode, the agent chooses to unmask a certain entry of $x$. 
When ready, the guesser may choose to predict the outcome $y$ from the unmasked features. 
When this is the case, the agent is rewarded according to the quality of the guesser's prediction. 
Therefore, it learns to select features that will allow the guesser to make an accurate guess.
During training, two separate optimization procedures take place, where both the guesser and the agent are being trained.
Our approach is depicted in Figure~\ref{fig:1}.

\subsubsection {The MDP}\label{sec:mdp}

Our MDP is defined as follows:
\begin{itemize}
\item State space: $\mathcal{S} = \mathbb{R}^{2d}$, where the first $d$ entries correspond to the patient's answers to the $d$ interview questions , and for $i=1,\ldots, d$, the $d+ i$ entry is set to 1 if question $i$ was chosen and 0 otherwise.
\item Action space:  $\mathcal{A} = \{1, 2,\ldots, d+1 \}$, where actions $1,\ldots,d$ refer to choosing the corresponding question and action $d+1$ refers to making a guess about the outcome variable.  
To prevent the agent from selecting the same feature more than once, we apply masking, as will be explained in Section~\ref{sec:add}.
\item State transition rules:
At the beginning of an episode, the initial state is simply a zero vector of length $2d$.
At each time step throughout the episode, if the action refers to asking  a new question ($1 \le a \le d$), state $s'$ is obtained from state $s$ by unmasking the $a$'th entry of $s$  (i.e., setting $s'_a = x_a$) and marking that question $a$ was asked (i.e., setting $s'_{a+d} = 1$).
if the agent chose to make a guess (i.e., $a=d+1$), the state remains unchanged and the episode terminates.
\item Reward function: For any question action  ($1 \le a \le d$), the reward is a small random number: $\mathcal{R}(s, a) = 0.1 \cdot \mbox{Unif}(0,1)$. 
For a guess action (i.e., $a = d+1$), the reward corresponds to the probability the environment guesser $G$ assigns to the true label,  $\mathcal{R}(s, d+1) = \mbox{Prob}(G(s) = y)$.
Observe that the fact that the reward function is parametrized by the guesser network, which is trained as well along with the agent, makes the MDP non-stationary, as during the course of training the guesser's weights change and correspondingly so does the reward function. 
This non-stationarity of the MDP deviates from the majority of recent RL works, which consider a stationary setting. 
To cope with the challenges the non-stationary setting introduces, we alternate between training of the guesser and of the agent, as will be explained in Section~\ref{sec:add}.
 
\end{itemize}

\subsubsection {The Environment}\label{sec:env}
As mentioned above, we augment our environment with a Guesser network, which is trained along with the RL agent. The guesser $G$ maps a state $s$ to $G(s)$, which is the probability assigned by the guesser to a positive outcome $p(y=1|s)$.
At the beginning of each episode, we reset the state so that only the age, gender and race features are visible and all other features are masked. 
At each step of the episode, an additional feature becomes visible, corresponding to the agent's chosen actions.
The episode terminates whenever the number of steps exceeds the pre-defined number of steps, or earlier, if the agent chooses to make a guess about the patient's outcome. Whenever the agent chooses to make a prediction,  the guesser network is called to predict the outcome from the current state (i.e., from the collection of all unmasked features). 
The probability that the guesser assigns to the correct class (which is known during training) uses as the reward which is passed on to the agent. 

\subsubsection {The Agent}\label{sec:agent}
We chose to use a DDQN agent.
In our experiments this model performed at the same level or even outperformed more sophisticated recent models such as PPO~\citep{schulman2017proximal}.

\subsubsection {Additional Design Choices for Performance Improvements}\label{sec:add}
Several implementational details allow to improve the performance of the algorithm. Below we briefly describe them.

\noindent \textbf{Oversampling}
Our dataset is highly unbalanced: less than 5 percent of the patients had positive outcome (died within four years from the date of filling the questionnaire).
To avoid bias toward the large class, in each training episode we sample a patient from one of the classes with equal probability (i.e., we over-sample the small class), so that roughly similar number of patients from each class are seen during training. 

\noindent \textbf{Alternating training}
To improve the training stability in our non-stationary setting, we trained the guesser and the agent intermittently, switching between them once in 1000 episodes. This way, during each such 1000 episodes period, when the agent is being trained, the guesser network remains fixed, so that the MDP is in fact (``locally'' ) stationary. 

\noindent \textbf{Pre-training the guesser network}
We pre-train the guesser G as a classifier, where all features are visible. 
When setting-up the environment, we initialized the guesser network using the parameters of the pre-trained guesser.

\noindent \textbf{Early Stopping}
Unlike typical RL works, we know the labels of the training data, which allows us to dedicate a portion of the data for validation set and apply an early stopping mechanism.
Specifically, every 1000 training episodes we run our agent on the validation set and record its AUC. Training stops when no significant improvements of AUC occurs. 
In inference, we use the model with the highest AUC on the validation set.

\noindent \textbf{Masking}
In order to ensure that the agent avoids selecting the same feature more than once, we apply a multiplicative mask to the agents' Q values so that Q values of features that were already selected are multiplied by zero and consequently will not be selected again 

\noindent \textbf{$\epsilon$-greedy sampling probabilities}
In order to explore new paths, it is a common practice to select the action that maximizes the $Q$ values in equations~\eqref{eq:loss} and~\eqref{eq:ddqn} with probability $1-\epsilon$, rather than with probability 1, and select a action uniformly at random with probability $\epsilon$. Usually practitioners use a time-decay policy for $\epsilon$.
Here, instead of using uniform probabilities for action sampling, we sample each action $a$ with probability which is proportional to the absolute correlation of the corresponding question with the target label over the training set.
This heuristic helps choosing actions which are more informative about the target with higher probability.

\noindent \textbf{Architectures}
Our best results were achieved using straightforward depth 3 MLP architectures for both the guesser and the Q network.
We also experimented with a more sophisticated design, where the state update mechanism is a Long-Short-Term-Memory (LSTM~\citep{hochreiter1997long}) cell. 
In this design the input to the LSTM cell at each time step is an embedding of the identity of selected feature, concatenated to the actual value of the feature.

\section {Experimental Results}\label{sec:experimental}

We begin this section by a visual demonstration of the feature selection process. We then report our primary results in predicting 4-year mortality from the national survey dataset, and comparing it to major off-the-shelf baselines.
We then provide  results of ablation studies, justifying our main design choices, and of experimenting with our approach under off-policy regime.
All results reported in this section are averaged over 5 identical trials with random splits of the dataset to train (67\%) and test (33\%) sets.

\subsection{Demonstration on Mnist}
\begin{figure}[t]
  \centering
   \includegraphics[width=.325\textwidth]{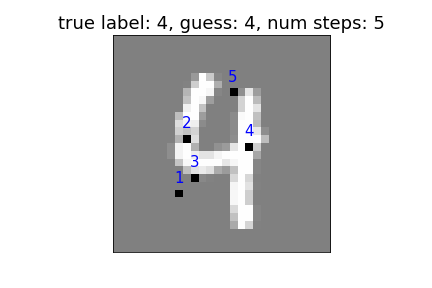}
   \includegraphics[width=.325\textwidth]{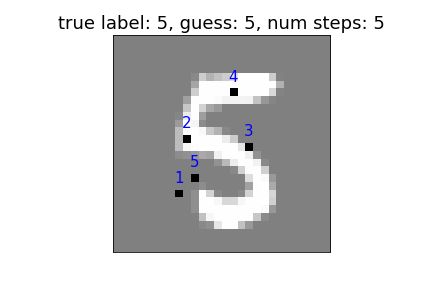}
   \includegraphics[width=.325\textwidth]{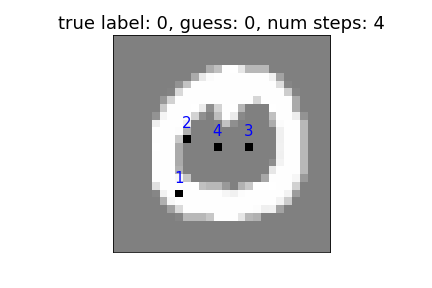}
   \includegraphics[width=.325\textwidth]{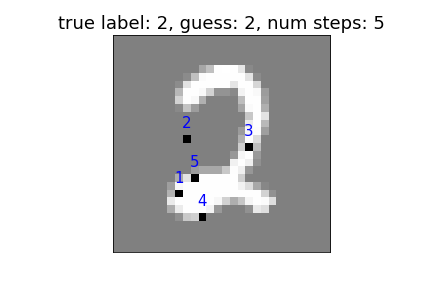}     
   \includegraphics[width=.325\textwidth]{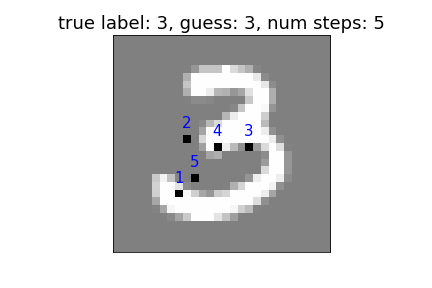}
   \includegraphics[width=.325\textwidth]{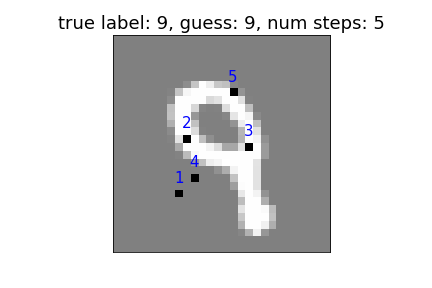}      
    \caption{Demonstration of the proposed approach on the Mnist handwritten digit dataset. Unmasked features appear in black. The order in which the features were selected is indicated in {\color{blue} blue}.}
    \label{fig:mnist}
\end{figure}

For the purpose of demonstration, we applied our approach on the Mnist handwritten digits dataset, where each pixel is a feature.  The goal is to predict the handwritten digits based on at most five pixels. The agent was able to correctly recognize the handwritten digit from at most 5 pixels 56.9\% of the time.
Figure~\ref{fig:mnist} shows examples of predictions made by the algorithm, and the corresponding selected features.

\subsection{Main Results}
Our goal is to obtain a good prediction for the outcome variable while considering a small number of features for each patient.
As baselines, we choose to compare our result to two off-the-shelf classifiers: a Decision Tree (DT) and XGBoost (XGB)~\citep{chen2016xgboost}.

A decision tree is a fundamental and widely-used machine learning algorithm. It has several known disadvantages, such as its greedy training procedure and its simplistic modeling of the feature space (axis-aligned rectangles), which often prevent it from performing on par with the state-of-the-art models. 
However, in many cases it is nevertheless a strong performer.
In our context, a DT has three attractive attributes which make it an appropriate baseline: first, specifying the depth of the tree, we can limit the number of features leading to each prediction made by the model. Second, different patients might correspond to different paths down the tree, so that different subsets of features may be used to obtain the predictions of different patients. Third, a DT is fully transparent, so that the predictions have high interpretability.

XGBoost is arguably considered as the state-of-the-art model for tabular data and is a popular choice by practitioners. Being an ensemble method, its interpretability is low, in the sense that it is difficult to provide a clear reasoning to the prediction made by the algorithm. 
Yet, given a trained model it is possible to obtain feature importance scores, describing how important each feature is to the predictions made by the model (see~\citep{lundberg2018consistent}, for example). 
In addition, we use the feature importance scores in order to reduce the number of features prior to applying our proposed approach.
The results reported in this manuscript were obtained by letting the agent select features out of the 50 most important features of a XGB model. We get similar results for the 100 most important features as well.
The list of these 50 features appears in Appendix~\ref{app:feat}.
 
Figure~\ref{fig:performance} shows the test AUC results of the proposed approach, comparing to DT and XGB. 
For every number of features $k$, the DT was developed up to depth $k$, the XGB model was trained on the subset of $k$ most important features of a full XGB model (trained on all features), and the RL agent was trained to choose $k-3$ features, as it was forced to select the age, gender and race features as starting point.
\begin{figure}[h!]
  \centering
   \includegraphics[width=1.0\textwidth]{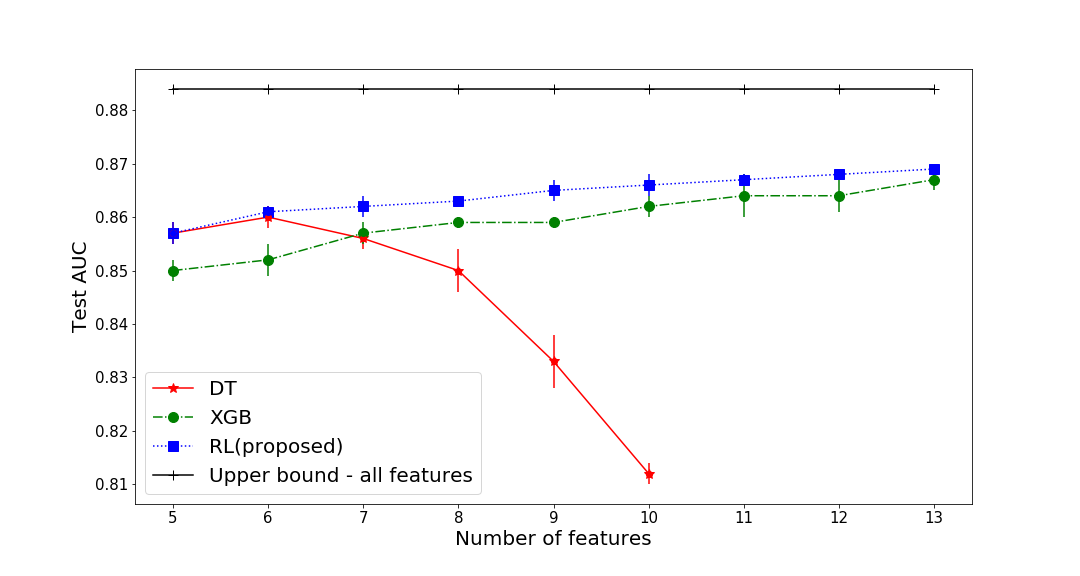}  
    \caption{Test AUC performance of DT, XGB and our proposed RL approach. The upper bound was obtained using all features; the same performance was achieved by both XGB and a MLP network.}
    \label{fig:performance}
\end{figure}
Our proposed RL approach consistently outperforms both the DT and the XGB models for all tested numbers of features. 
Moreover, the RL models improve monotonically as more features are allowed to be selected, as is also the case for the XGB models. The DT models, however, start to overfit for more than 6 features.

The advantage of the proposed approach over XGB manifests not only in terms of prediction accuracy, but also in terms of interpretability, through fact that one gets to observe the sequence of unmasked features leading to the each prediction. 
the predictions of our approach are provided in Appendix~\ref{app:examples}.

Figure~\ref{fig:cases} shows two case studies of the model predictions for patients from the test set.
\begin{figure}[h!]
  \centering
   \includegraphics[width=0.49\textwidth]{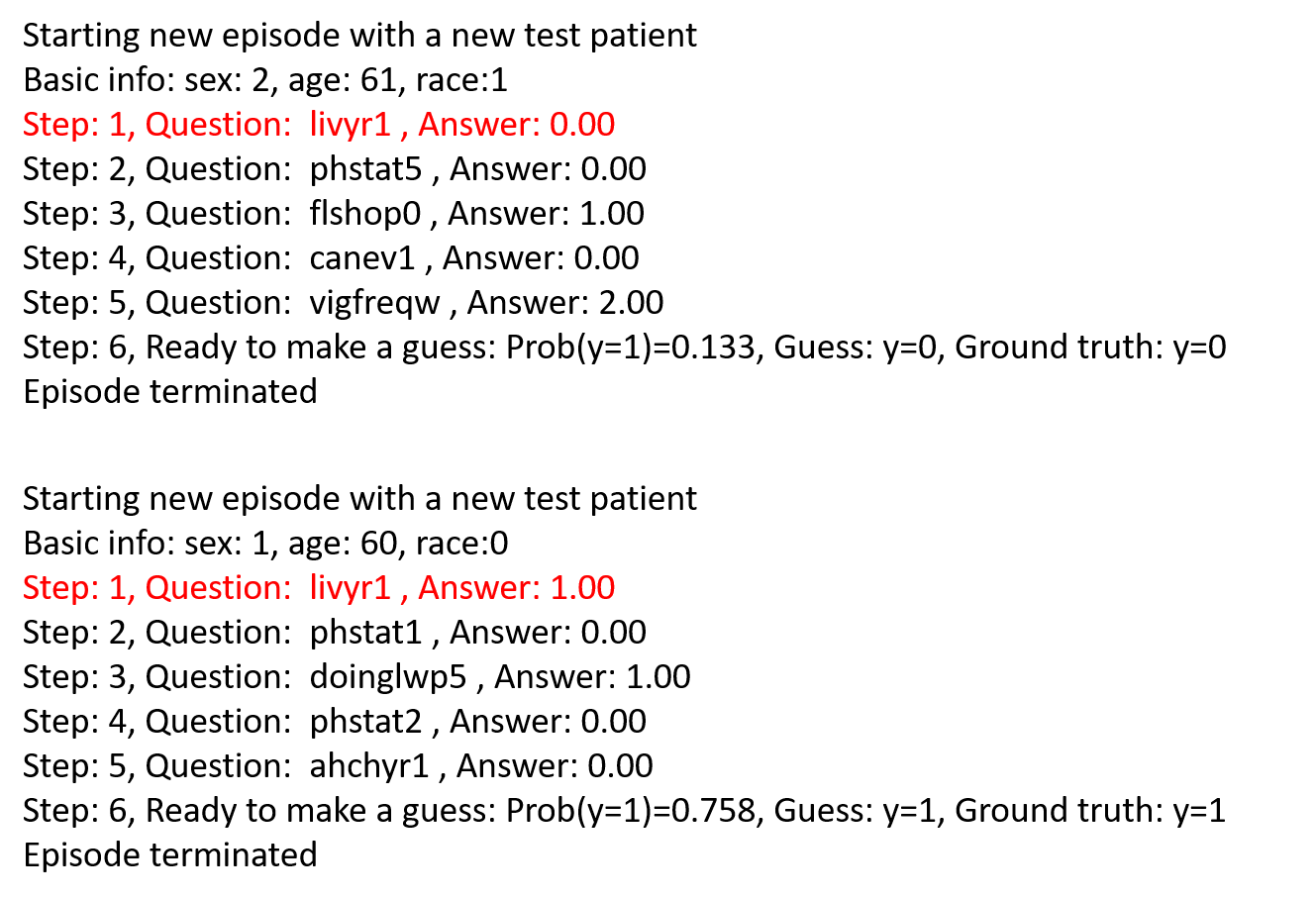}  
   \includegraphics[width=0.49\textwidth]{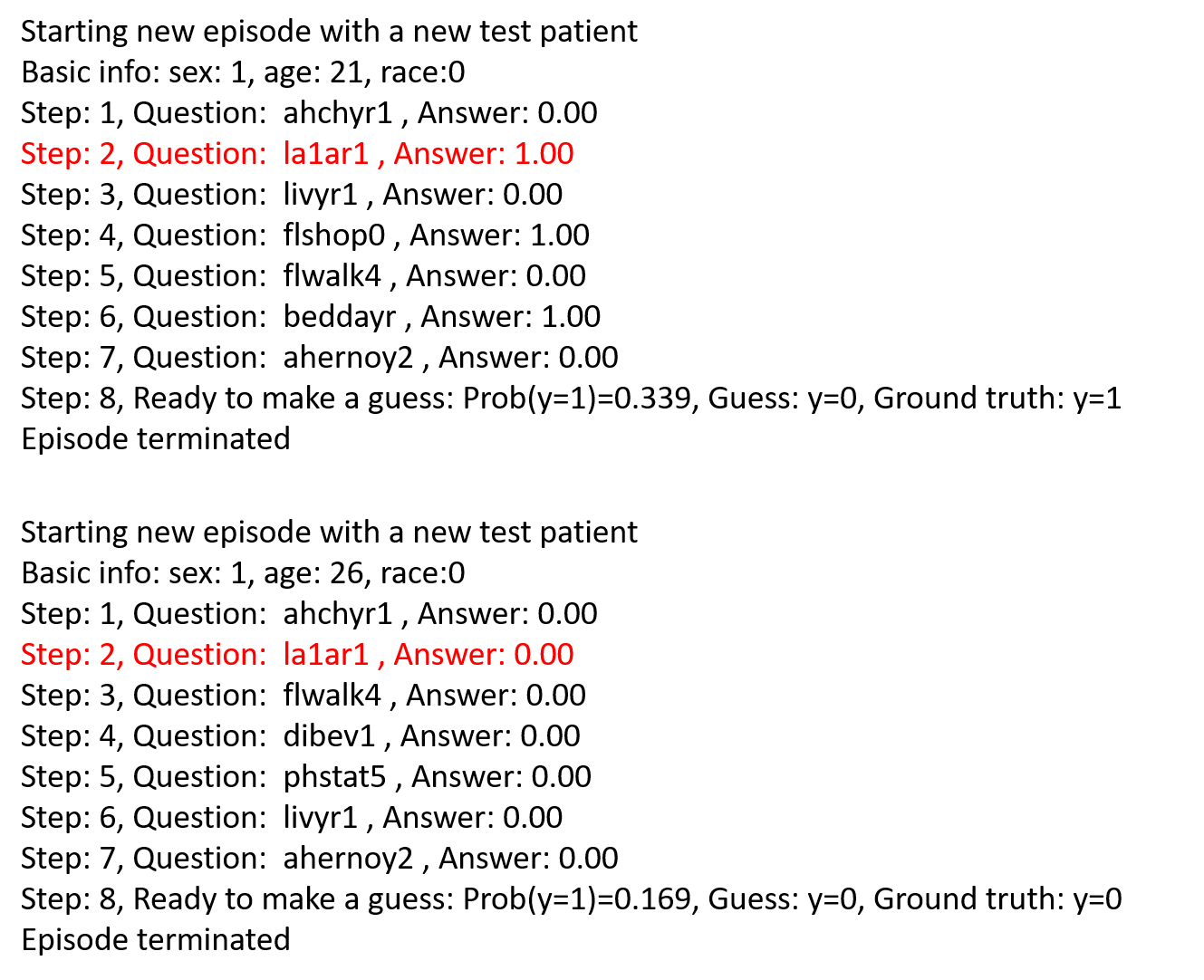}  
    \caption{Two case studies of the agent behavior.}
    \label{fig:cases}
\end{figure}
On the left hand side of Figure~\ref{fig:cases} the agent acts on the data of two patients from the same race, having similar age and different sex.
For both patients the agent chooses to unmask a feature containing information about the patients' liver condition (marked in red). The bottom patient had a liver condition while the top patient did not have such condition. From the second step and on the agent chooses to unmask different features for each patient, leading to a prediction of low probability for mortality for the top patient (whose unmasked features revealed is not in a poor physical condition, did not have cancer and does not need special equipment to go out), whereas the bottom patient (whose unmasked features revealed was not in a good physical condition and did not go to work last week) is assigned a high probability for 4-year mortality.

In the second case study, on the right hand side of Figure~\ref{fig:cases}, the agents acts on two young patients of the same race and sex. The second unmasked feature reveals that the top patient is limited in some way, while the bottom one is not. This leads to different unmasked features for each patient, revealing that the top patient also has necessity for special equipment and had bed days during the past 12 months, while for the bottom patient no potential negative medical conditions are recognized.  The episodes end with a 4-year mortality probability assignment which is twice as high for the top patient (who indeed died within 4 years of the questionnaire).

\subsection{Off-Policy Experiments}
Off-policy learning is an important area in RL. It corresponds to cases where the states the agent observes are not a consequence of the agent's policy. 
In cases like this there might be a decrease in the performance of the agent, as it was not trained on such states.
$Q$ learning is an off-policy learning method, as it updates the Q function independently of the policy it currently follows, .i.e., the updates are based on tuples $(s_t, a_t, r_t, s_{t+1})$ from past versions of the policy.
Being an off-policy learner, DDQN handles such cases by design.
To verify the performance of our proposed approach is stable under an off-policy regime, we considered a case where some features are given to us ``for free'', i.e., along with the age, gender and race of the patient we may also observe additional features, without the agent explicitly choosing to unmask these features.
In order to investigate the performance of our proposed approach under such a scenario, we train the guesser and agent as usual, but modify our inference procedure, so when the environment restarts the state at the beginning of any episode, along with unmasking the age, gender and race features, it also unmasks a randomly chosen feature (selected randomly for each new test patient).
Applying this test procedure for $k=10$ features, we observed that the AUC over the test set decreased from 0.865 to 0.862. While a slight decrease is somewhat expected, the decrease is relatively minor, and the model seem to perform roughly on the same level as before. 

\subsection{Ablation studies}
In this section we investigate the contribution of the oversampling, guesser pre-training and alternation of the training of the guesser and Q network. The results for $k=10$ features appear in Table~\ref{tab:ablation}.

\begin{table}[h!]
\centering
\begin{tabular}{|| l | c ||} 
 \hline
 Configuration & Test AUC \\ [0.5ex] 
 \hline\hline
  Full approach (proposed)       & 0.865 (0.003)   \\ [0.5ex] 
  No guesser pre-training         & 0.856 (0.003)   \\ [0.5ex]
  No oversampling                  & 0.855 (0.002)   \\ [0.5ex] 
  No alternation                      & 0.834 (0.002)   \\ [0.5ex]  
 \hline
\end{tabular}
\caption{Ablation studies.}
\label{tab:ablation}
\end{table}
As can be seen, absence of any of the three elements causes a decrease in performance, comparing to the full approach.

\subsection{Question Embedding}
\begin{figure}[t]
  \centering
   \includegraphics[width=1.0\textwidth]{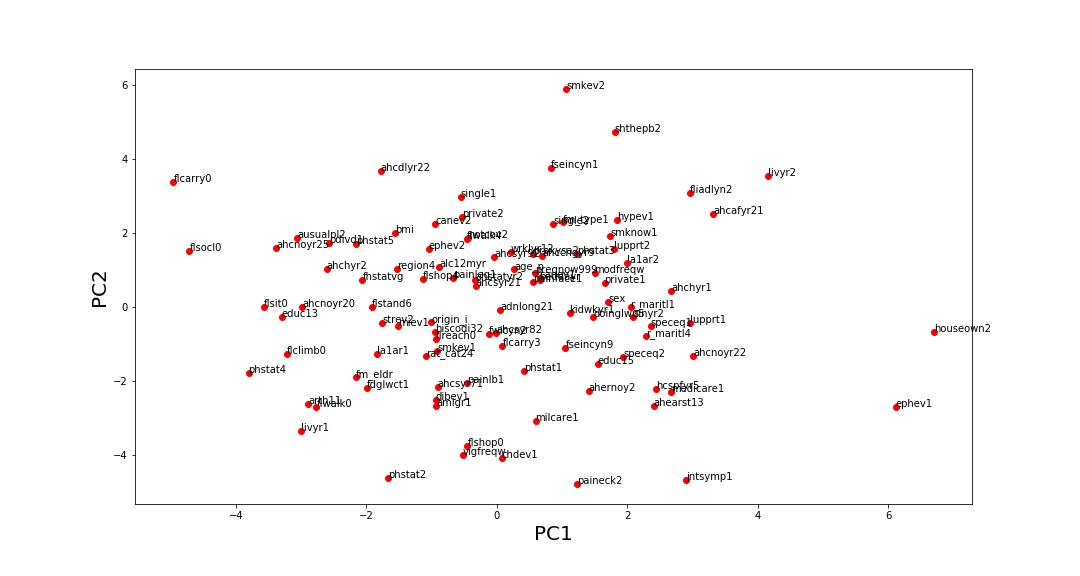}  
    \caption{Question embedding using LSTM architecture. The two plotted dimensions are the first two principal axes of the 64-dimensional embedding.}
    \label{fig:embedding}
\end{figure}

Figure~\ref{fig:embedding} shows the embedding of the questions, obtained from training our proposed approach using the LSTM architecture.
Interestingly, the plot manifests several intuitive relations between pairs of features. For example, the feature vectors of \textit{la1ar1} (limited in any way) and \textit{la1ar2} (not limited in any way) are in opposite directions, and so are \textit{phstat1} (excellent health status) and  \textit{phstat5} (poor health status), as well as these of \textit{livyr1} (told to have liver condition)  and \textit{livyr2} (where not told).
On the other hand, the feature vectors of  \textit{hiscodi32} and \textit{origin\_i}, both correspond to Hispanic origin, are close.

\subsection{Technical details}
We used the sklearn implementation of DT, where for $k$ features we built a full binary tree of depth $k$. We were not able to obtain better results using pruning.
For XGBoost we used the python \textit{xgboost} package, with ensemble size of 100. For each number $k$ of features we manually tuned the tree depth to achieve the best performance.
For both DT and XGB models we used class weights for training, such that the weight for each class was inversely proportional to its relative size.

For the proposed RL approach, we used the same hyperparameter setting for all numbers of features.
The guesser architecture was a multi-layer perceptron (MLP) architecture, with three hidden layers of 250 PReLU units each and a softmax output layer.
The Q network architecture had two hidden layers of 128 ReLU units each and sigmoid output layer.
Weight penalty was added to the DQN loss.
For both networks we used a learning rate decay policy, with initial value of $1e-4$ and division by 10 every 17500 training episodes, with minimal learning rate of $1e-6$.
We set the reward decay factor $\gamma$ to 0.95.

\section {Conclusions}\label{sec:conclusions}

In this manuscript we proposed a reinforcement learning-based approach for adaptive feature selection and applied it to a  national survey dataset, where the goal is to predict 4-year mortality of patients.
We demonstrated that our approach outperforms standard baseline models for the same task, while also being more interpretable than its closest competitor models.
In the future we plan to use this approach as a basis for recommendation system and extend it to other types of medical data, such as images and natural language.
In addition, we plan  to incorporate feature costs and non-symmetric error costs into the model, by modifications of the loss function.

\bibliography{questionnaire}

\begin{thebibliography}{}

\bibitem[\protect\citeauthoryear{Bal{\i}n \bgroup \em et al.\egroup
  }{2019}]{balin2019concrete}
Muhammed~Fatih Bal{\i}n, Abubakar Abid, and James Zou.
\newblock Concrete autoencoders: Differentiable feature selection and
  reconstruction.
\newblock In {\em International Conference on Machine Learning}, pages
  444--453, 2019.

\bibitem[\protect\citeauthoryear{Blewett \bgroup \em et al.\egroup
  }{2016}]{blewett2016ipums}
Lynn~A Blewett, Julia~A Rivera~Drew, Risa Griffin, Miram~L King, and Kari
  Williams.
\newblock Ipums health surveys: National health interview survey, version 6.2.
\newblock {\em Minneapolis: University of Minnesota}, 10:D070, 2016.

\bibitem[\protect\citeauthoryear{Chen and Guestrin}{2016}]{chen2016xgboost}
Tianqi Chen and Carlos Guestrin.
\newblock Xgboost: A scalable tree boosting system.
\newblock In {\em Proceedings of the 22nd acm sigkdd international conference
  on knowledge discovery and data mining}, pages 785--794, 2016.

\bibitem[\protect\citeauthoryear{Chen \bgroup \em et al.\egroup
  }{2018}]{chen2018learning}
Yihong Chen, Bei Chen, Xuguang Duan, Jian-Guang Lou, Yue Wang, Wenwu Zhu, and
  Yong Cao.
\newblock Learning-to-ask: Knowledge acquisition via 20 questions.
\newblock In {\em Proceedings of the 24th ACM SIGKDD International Conference
  on Knowledge Discovery \& Data Mining}, pages 1216--1225, 2018.

\bibitem[\protect\citeauthoryear{Ernst \bgroup \em et al.\egroup
  }{2005}]{ernst2005tree}
Damien Ernst, Pierre Geurts, and Louis Wehenkel.
\newblock Tree-based batch mode reinforcement learning.
\newblock {\em Journal of Machine Learning Research}, 6(Apr):503--556, 2005.

\bibitem[\protect\citeauthoryear{Hochreiter and
  Schmidhuber}{1997}]{hochreiter1997long}
Sepp Hochreiter and J{\"u}rgen Schmidhuber.
\newblock Long short-term memory.
\newblock {\em Neural computation}, 9(8):1735--1780, 1997.

\bibitem[\protect\citeauthoryear{Hu \bgroup \em et al.\egroup
  }{2018}]{hu2018playing}
Huang Hu, Xianchao Wu, Bingfeng Luo, Chongyang Tao, Can Xu, Wei Wu, and Zhan
  Chen.
\newblock Playing 20 question game with policy-based reinforcement learning.
\newblock {\em arXiv preprint arXiv:1808.07645}, 2018.

\bibitem[\protect\citeauthoryear{Li \bgroup \em et al.\egroup
  }{2016}]{li2016deep}
Yifeng Li, Chih-Yu Chen, and Wyeth~W Wasserman.
\newblock Deep feature selection: theory and application to identify enhancers
  and promoters.
\newblock {\em Journal of Computational Biology}, 23(5):322--336, 2016.

\bibitem[\protect\citeauthoryear{Li \bgroup \em et al.\egroup
  }{2018}]{li2018feature}
Jundong Li, Kewei Cheng, Suhang Wang, Fred Morstatter, Robert~P Trevino,
  Jiliang Tang, and Huan Liu.
\newblock Feature selection: A data perspective.
\newblock {\em ACM Computing Surveys (CSUR)}, 50(6):94, 2018.

\bibitem[\protect\citeauthoryear{Louizos \bgroup \em et al.\egroup
  }{2017}]{louizos2017learning}
Christos Louizos, Max Welling, and Diederik~P Kingma.
\newblock Learning sparse neural networks through $ l\_0 $ regularization.
\newblock {\em arXiv preprint arXiv:1712.01312}, 2017.

\bibitem[\protect\citeauthoryear{Lundberg \bgroup \em et al.\egroup
  }{2018}]{lundberg2018consistent}
Scott~M Lundberg, Gabriel~G Erion, and Su-In Lee.
\newblock Consistent individualized feature attribution for tree ensembles.
\newblock {\em arXiv preprint arXiv:1802.03888}, 2018.

\bibitem[\protect\citeauthoryear{Mnih \bgroup \em et al.\egroup
  }{2015}]{mnih2015human}
Volodymyr Mnih, Koray Kavukcuoglu, David Silver, Andrei~A Rusu, Joel Veness,
  Marc~G Bellemare, Alex Graves, Martin Riedmiller, Andreas~K Fidjeland, Georg
  Ostrovski, et~al.
\newblock Human-level control through deep reinforcement learning.
\newblock {\em Nature}, 518(7540):529--533, 2015.

\bibitem[\protect\citeauthoryear{Roy \bgroup \em et al.\egroup
  }{2015}]{roy2015feature}
Debaditya Roy, K~Sri~Rama Murty, and C~Krishna Mohan.
\newblock Feature selection using deep neural networks.
\newblock In {\em 2015 International Joint Conference on Neural Networks
  (IJCNN)}, pages 1--6. IEEE, 2015.

\bibitem[\protect\citeauthoryear{Schulman \bgroup \em et al.\egroup
  }{2017}]{schulman2017proximal}
John Schulman, Filip Wolski, Prafulla Dhariwal, Alec Radford, and Oleg Klimov.
\newblock Proximal policy optimization algorithms.
\newblock {\em arXiv preprint arXiv:1707.06347}, 2017.

\bibitem[\protect\citeauthoryear{Silver \bgroup \em et al.\egroup
  }{2016}]{silver2016mastering}
David Silver, Aja Huang, Chris~J Maddison, Arthur Guez, Laurent Sifre, George
  Van Den~Driessche, Julian Schrittwieser, Ioannis Antonoglou, Veda
  Panneershelvam, Marc Lanctot, et~al.
\newblock Mastering the game of go with deep neural networks and tree search.
\newblock {\em nature}, 529(7587):484, 2016.

\bibitem[\protect\citeauthoryear{Tibshirani}{1996}]{tibshirani1996regression}
Robert Tibshirani.
\newblock Regression shrinkage and selection via the lasso.
\newblock {\em Journal of the Royal Statistical Society: Series B
  (Methodological)}, 58(1):267--288, 1996.

\bibitem[\protect\citeauthoryear{Van~Hasselt \bgroup \em et al.\egroup
  }{2016}]{van2016deep}
Hado Van~Hasselt, Arthur Guez, and David Silver.
\newblock Deep reinforcement learning with double q-learning.
\newblock In {\em Thirtieth AAAI conference on artificial intelligence}, 2016.

\bibitem[\protect\citeauthoryear{Watkins and Dayan}{1992}]{watkins1992q}
Christopher~JCH Watkins and Peter Dayan.
\newblock Q-learning.
\newblock {\em Machine learning}, 8(3-4):279--292, 1992.

\bibitem[\protect\citeauthoryear{Yamada \bgroup \em et al.\egroup
  }{2018}]{yamada2018deep}
Yutaro Yamada, Ofir Lindenbaum, Sahand Negahban, and Yuval Kluger.
\newblock Deep supervised feature selection using stochastic gates.
\newblock {\em arXiv preprint arXiv:1810.04247}, 2018.

\bibitem[\protect\citeauthoryear{Zhao and Eskenazi}{2016}]{zhao2016towards}
Tiancheng Zhao and Maxine Eskenazi.
\newblock Towards end-to-end learning for dialog state tracking and management
  using deep reinforcement learning.
\newblock {\em arXiv preprint arXiv:1606.02560}, 2016.

\bibitem[\protect\citeauthoryear{Zhao \bgroup \em et al.\egroup
  }{2015}]{zhao2015heterogeneous}
Lei Zhao, Qinghua Hu, and Wenwu Wang.
\newblock Heterogeneous feature selection with multi-modal deep neural networks
  and sparse group lasso.
\newblock {\em IEEE Transactions on Multimedia}, 17(11):1936--1948, 2015.

\bibitem[\protect\citeauthoryear{Zheng \bgroup \em et al.\egroup
  }{2018}]{zheng2018learning}
Zeyu Zheng, Junhyuk Oh, and Satinder Singh.
\newblock On learning intrinsic rewards for policy gradient methods.
\newblock In {\em Advances in Neural Information Processing Systems}, pages
  4644--4654, 2018.
  
\bibitem[\protect\citeauthoryear{Sutton and
  Barto}{2018}]{sutton2018reinforcement}
Richard S Sutton and Andrew G Barto.
\newblock Reinforcement learning: An introduction.
\newblock {\em MIT press}, 2018.  

\end{thebibliography}
\bibliographystyle{named}
\appendix

\section {NHIS-NDI 2002-2011 Data Pre-processing Workflow}\label{sec:dpw}

\noindent {\bf Step 1}
\begin{enumerate}
\item We downloaded the publicly available data files for NHIS from the CDC website for years 2002 - 2011. We merged data from 3 separate files for each individual year - 1) sample adult file, 2) family file, and 3) person file. \url{a.	https://www.cdc.gov/nchs/nhis/data-questionnaires-documentation.htm}
\item We added 3 variables from household files for years 2002-2004 (month of interview [2002, 2003], year of interview [2002, 2003], and region [2004]) since they were in the household files for those years but in the person file for years 2005-2011. We merged the 10 years of complete NHIS data for years 2002-2011.
\item We obtained the variance estimation for the entire 10 year pooled cohort from the Integrated Public Use Microdata Series \url{https://nhis.ipums.org/nhis/}~\citep{blewett2016ipums}.
\item We merged the NHIS data set with Public-use Linked Mortality Files that extract mortality information from the National Death Index \url{https://www.cdc.gov/nchs/data-linkage/mortality-public.htm}.
\item The resulting dataset contained $n=282,001$ observations and $d=2,360$ variables.
\end{enumerate}

\noindent {\bf Step 2}
\begin{enumerate}
\item We dropped the observations without mortality information ($n=12,905$). 
\item We created 3 new variables for defining the outcome – 1) interview quarter (iv-qtr), 2) interval to death (int-death), and 3) death within 4 years (dead-4y).
\item The resulting dataset contained $n=269,096$ observations and $d=2,363$ variables.
\end{enumerate}

\noindent {\bf Step 3}
\begin{enumerate}
\item  Using the data questionnaire documentations for each year, we listed all the variables that had their name changed across the years, keeping the name of their most recent appearance. 
\item Kept the variables that were consistent across the years.
\item  The resulting dataset contained $n=269,096$ observations and $d1,022$ variables.
\end{enumerate}

\noindent {\bf Step 4}
\begin{enumerate}
\item From the list of all consistent variables, each variable was reviewed by 2 different investigators independently (SM, DM, AA, or CC) and flagged as parent or daughter variable based on the data questionnaires documents and type of question, and kept only the parent variables. Altogether 264 variables were kept.
\item Identified variables with $>80$\% missing values not from the NDI variables (like cause of death) and dropped them. 8 variables were dropped.
\item We dropped variables that had their information contained under other variables and household/family identifiers. 45 variables were dropped.
\item  The resulting dataset contained $n=269,096$ observations and $d211$ variables.
\end{enumerate}

\noindent {\bf Step 5}
\begin{enumerate}
\item We re-coded variables for analysis using the following guidelines:
	\begin{itemize}
	\item Categorical variables: We replaced missing values as a separate “999” category for 19 variables. Note: we reduced the number of categories for the following 4 variables by collapsing their values - income ratio (rat-cat2), education (educ1), usual place of care (ausualpl), and family structure (fm-strp).
	\item Numeric  variables: 	For numeric variables with values of 97, 98, and 99 (refused, don’t know, missing) (9 variables), we created new variables for each of 97, 98, 99 categories (e.g. varx-97), and replaced those values in the original variable as missing (.). 
	\end{itemize}
\item Median single-value imputation for missing values for the continuous variables (10 variables).
\item The resulting dataset contained $n=269,096$ observations and $d=242$ variables.
\end{enumerate}

\noindent {\bf Step 6}
\begin{enumerate}
\item One-hot encoding for categorical interview variables (156 variables) in R. 867 One-hot encoded interview variables were generated.
\item \item The resulting dataset contained $n=269,096$ observations, $d=932$ interview variables, 14 identifiers and 9 outcome variables, out of which we considered the 4 year mortality variable.
\end{enumerate}

\section {Input features}\label{app:feat}
Table~\ref{tab:features} shows the set of 50 most important features of the XGB model, trained on the full feature set. In all experiments in this manuscript features were selected from this set.

\begin{table}[h!]
\tiny
\centering
\begin{tabular}{|| l | c ||} 
 \hline
  Feature name & Meaning \\ [0.5ex] 
 \hline\hline
medicare1   & Medicare coverage recode \\ [0.5ex] 
la1ar1          & Any limitation - all persons, all conditions \\ [0.5ex] 
flwalk0         & How difficult to walk 1/4 mile without special equipment \\ [0.5ex] 
age-p           & Age \\ [0.5ex] 
flclimb0         & How difficult to climb 10 steps without special equipment \\ [0.5ex] 
doinglwp5     & What was - - doing last week \\ [0.5ex] 
la1ar2          & Any limitation - all persons, all conditions \\ [0.5ex] 
flcarry0        & How difficult to lift/carry 10 lbs without special equipment \\ [0.5ex] 
wrklyr12       & Work for pay last year \\ [0.5ex] 
pregnow999 & Currently pregnant \\ [0.5ex] 
smkev1         & Ever smoked 100 cigarettes \\ [0.5ex] 
lupprt1          & Lost all upper and lower natural teeth \\ [0.5ex] 
phstat5         & Reported health status \\ [0.5ex] 
speceq2        & Have health problem that requires special equipment \\ [0.5ex] 
flshop0         & How difficult to go out to events without special equipment \\ [0.5ex] 
flwalk4          & How difficult to walk 1/4 mile without special equipment \\ [0.5ex] 
fliadlyn2        & Any family member need help with an IADL \\ [0.5ex] 
smkev2         & Ever smoked 100 cigarettes \\ [0.5ex] 
educ15         & Highest level of school completed \\ [0.5ex] 
phstat4        & Reported health status \\ [0.5ex] 
eligpwic        & Anyone age-eligible for the WIC program \\ [0.5ex] 
canev1        & Ever told by a doctor you had cancer \\ [0.5ex] 
adnlong21   & Time since last saw a dentist \\ [0.5ex] 
vigfreqw      & Freq vigorous activity (times per wk) \\ [0.5ex] 
sex              & Sex \\ [0.5ex] 
livyr2           & Told you had liver condition, past 12 m \\ [0.5ex] 
private2      & Private health insurance recode \\ [0.5ex] 
ahchyr1      & Received home care from health professional, past 12 m \\ [0.5ex] 
ahcsyr71     & Seen/talked to mental health professional, past 12 m \\ [0.5ex] 
smknow1     & Smoke freq: everyday/some days/not at all \\ [0.5ex] 
origin-i         & Hispanic Ethnicity \\ [0.5ex] 
dibev1         & Ever been told that you have diabetes \\ [0.5ex] 
ephev1       & Ever been told you had emphysema \\ [0.5ex] 
miev1          & Ever been told you had a heart attack \\ [0.5ex] 
kidwkyr2     & Told you had weak/failing kidneys, 12 m \\ [0.5ex] 
phstat1      & Reported health status \\ [0.5ex] 
flsocl0        & How difficult to participate in social activities without speci \\ [0.5ex] 
phstat2      & Reported health status \\ [0.5ex] 
ahchyr2     & Received home care from health professional, past 12 m \\ [0.5ex] 
hiscodi32   & Race/ethnicity recode \\ [0.5ex] 
livyr1         & Told you had liver condition, past 12 m \\ [0.5ex] 
bmi            & Body Mass Index (BMI) \\ [0.5ex] 
amigr2       & Had severe headache/migraine, past 3 m \\ [0.5ex] 
rat-cat24  & Ratio of family income to the poverty threshold \\ [0.5ex] 
jntsymp1  & Symptoms of joint pain/aching/stiffness past 30 d \\ [0.5ex] 
houseown2 & Home tenure status \\ [0.5ex] 
doinglwp1  & What was - - doing last week \\ [0.5ex] 
beddayr  & Number of bed days, past 12 months \\ [0.5ex] 
ahernoy2 & times in ER/ED, past 12 m \\ [0.5ex] 
proxysa2 & Sample adult status \\ [0.5ex] 
 \hline
\end{tabular}
\caption{The pool of 50 most important features of an XGBoost model, out of which the methods selected features.}
\label{tab:features}
\end{table}

\section {Examples}\label{app:examples}

\noindent Starting new episode with a new test patient \\ [0.5ex] 
Basic info: sex: 2, age: 85, race:0 \\ [0.5ex] 
Step: 1, Question:  la1ar2 , Answer: 0.00 \\ [0.5ex] 
Step: 2, Ready to make a guess: Prob(y=1)=0.874, Guess: y=1, Ground truth: y=1 \\ [0.5ex] 
Episode terminated \\ [0.5ex] 

\noindent Starting new episode with a new test patient \\ [0.5ex] 
Basic info: sex: 2, age: 24, race:0 \\ [0.5ex] 
Step: 1, Question:  la1ar2 , Answer: 1.00 \\ [0.5ex] 
Step: 2, Question:  proxysa2 , Answer: 1.00 \\ [0.5ex] 
Step: 3, Ready to make a guess: Prob(y=1)=0.147, Guess: y=0, Ground truth: y=0 \\ [0.5ex] 
Episode terminated \\ [0.5ex] 

\noindent Starting new episode with a new test patient \\ [0.5ex] 
Basic info: sex: 2, age: 67, race:1 \\ [0.5ex] 
Step: 1, Question:  la1ar2 , Answer: 1.00 \\ [0.5ex] 
Step: 2, Question:  ephev1 , Answer: 0.00 \\ [0.5ex] 
Step: 3, Question:  dibev1 , Answer: 0.00 \\ [0.5ex] 
Step: 4, Question:  kidwkyr2 , Answer: 1.00 \\ [0.5ex] 
Step: 5, Question:  proxysa2 , Answer: 1.00 \\ [0.5ex] 
Step: 6, Ready to make a guess: Prob(y=1)=0.299, Guess: y=0, Ground truth: y=1 \\ [0.5ex] 
Episode terminated \\ [0.5ex] 

\noindent Starting new episode with a new test patient \\ [0.5ex] 
Basic info: sex: 2, age: 20, race:0 \\ [0.5ex] 
Step: 1, Question:  la1ar2 , Answer: 1.00 \\ [0.5ex] 
Step: 2, Question:  proxysa2 , Answer: 1.00 \\ [0.5ex] 
Step: 3, Ready to make a guess: Prob(y=1)=0.143, Guess: y=0, Ground truth: y=0 \\ [0.5ex] 
Episode terminated \\ [0.5ex] 

\noindent Starting new episode with a new test patient \\ [0.5ex] 
Basic info: sex: 2, age: 48, race:0 \\ [0.5ex] 
Step: 1, Question:  smkev1 , Answer: 1.00 \\ [0.5ex] 
Step: 2, Question:  phstat1 , Answer: 0.00 \\ [0.5ex] 
Step: 3, Question:  kidwkyr2 , Answer: 1.00 \\ [0.5ex] 
Step: 4, Question:  flsocl0 , Answer: 0.00 \\ [0.5ex] 
Step: 5, Question:  ahernoy2 , Answer: 3.00 \\ [0.5ex] 
Step: 6, Question:  jntsymp1 , Answer: 1.00 \\ [0.5ex] 
Step: 7, Ready to make a guess: Prob(y=1)=0.437, Guess: y=0, Ground truth: y=0 \\ [0.5ex] 
Episode terminated \\ [0.5ex] 

\noindent Starting new episode with a new test patient \\ [0.5ex] 
Basic info: sex: 1, age: 83, race:1 \\ [0.5ex] 
Step: 1, Ready to make a guess: Prob(y=1)=0.906, Guess: y=1, Ground truth: y=1 \\ [0.5ex] 
Episode terminated \\ [0.5ex] 

\noindent Starting new episode with a new test patient \\ [0.5ex] 
Basic info: sex: 2, age: 27, race:1 \\ [0.5ex] 
Step: 1, Ready to make a guess: Prob(y=1)=0.082, Guess: y=0, Ground truth: y=0 \\ [0.5ex] 
Episode terminated \\ [0.5ex] 

\noindent Starting new episode with a new test patient \\ [0.5ex] 
Basic info: sex: 2, age: 64, race:1 \\ [0.5ex] 
Step: 1, Question:  smkev1 , Answer: 1.00 \\ [0.5ex] 
Step: 2, Question:  kidwkyr2 , Answer: 1.00 \\ [0.5ex] 
Step: 3, Question:  phstat1 , Answer: 0.00 \\ [0.5ex] 
Step: 4, Question:  phstat2 , Answer: 0.00 \\ [0.5ex] 
Step: 5, Question:  proxysa2 , Answer: 0.00 \\ [0.5ex] 
Step: 6, Question:  flwalk0 , Answer: 0.00 \\ [0.5ex] 
Step: 7, Ready to make a guess: Prob(y=1)=0.817, Guess: y=1, Ground truth: y=1 \\ [0.5ex] 
Episode terminated \\ [0.5ex] 

\noindent Starting new episode with a new test patient \\ [0.5ex] 
Basic info: sex: 2, age: 51, race:1 \\ [0.5ex] 
Step: 1, Question:  smkev1 , Answer: 0.00 \\ [0.5ex] 
Step: 2, Question:  phstat1 , Answer: 1.00 \\ [0.5ex] 
Step: 3, Question:  kidwkyr2 , Answer: 1.00 \\ [0.5ex] 
Step: 4, Question:  houseown2 , Answer: 0.00 \\ [0.5ex] 
Step: 5, Ready to make a guess: Prob(y=1)=0.099, Guess: y=0, Ground truth: y=0 \\ [0.5ex] 
Episode terminated \\ [0.5ex] 

\noindent Starting new episode with a new test patient \\ [0.5ex] 
Basic info: sex: 2, age: 55, race:1 \\ [0.5ex] 
Step: 1, Question:  smkev1 , Answer: 0.00 \\ [0.5ex] 
Step: 2, Question:  phstat1 , Answer: 0.00 \\ [0.5ex] 
Step: 3, Question:  kidwkyr2 , Answer: 1.00 \\ [0.5ex] 
Step: 4, Question:  private2 , Answer: 0.00 \\ [0.5ex] 
Step: 5, Question:  livyr2 , Answer: 1.00 \\ [0.5ex] 
Step: 6, Question:  flwalk0 , Answer: 0.00 \\ [0.5ex] 
Step: 7, Ready to make a guess: Prob(y=1)=0.282, Guess: y=0, Ground truth: y=1 \\ [0.5ex] 
Episode terminated \\ [0.5ex] 

\noindent Starting new episode with a new test patient \\ [0.5ex] 
Basic info: sex: 2, age: 38, race:0 \\ [0.5ex] 
Step: 1, Question:  smkev1 , Answer: 1.00 \\ [0.5ex] 
Step: 2, Question:  phstat1 , Answer: 0.00 \\ [0.5ex] 
Step: 3, Question:  kidwkyr2 , Answer: 1.00 \\ [0.5ex] 
Step: 4, Question:  flsocl0 , Answer: 1.00 \\ [0.5ex] 
Step: 5, Question:  fliadlyn2 , Answer: 1.00 \\ [0.5ex] 
Step: 6, Question:  jntsymp1 , Answer: 0.00 \\ [0.5ex] 
Step: 7, Ready to make a guess: Prob(y=1)=0.257, Guess: y=0, Ground truth: y=0 \\ [0.5ex] 
Episode terminated \\ [0.5ex] 

Starting new episode with a new test patient \\ [0.5ex] 
Basic info: sex: 1, age: 65, race:1 \\ [0.5ex] 
Step: 1, Question:  smkev1 , Answer: 1.00 \\ [0.5ex] 
Step: 2, Question:  phstat1 , Answer: 0.00 \\ [0.5ex] 
Step: 3, Question:  kidwkyr2 , Answer: 1.00 \\ [0.5ex] 
Step: 4, Question:  flshop0 , Answer: 1.00 \\ [0.5ex] 
Step: 5, Question:  dibev1 , Answer: 0.00 \\ [0.5ex] 
Step: 6, Question:  amigr2 , Answer: 1.00 \\ [0.5ex] 
Step: 7, Ready to make a guess: Prob(y=1)=0.737, Guess: y=1, Ground truth: y=1 \\ [0.5ex] 
Episode terminated \\ [0.5ex] 

\noindent Starting new episode with a new test patient \\ [0.5ex] 
Basic info: sex: 2, age: 40, race:1 \\ [0.5ex] 
Step: 1, Question:  smkev1 , Answer: 1.00 \\ [0.5ex] 
Step: 2, Question:  phstat1 , Answer: 0.00 \\ [0.5ex] 
Step: 3, Question:  kidwkyr2 , Answer: 1.00 \\ [0.5ex] 
Step: 4, Question:  eligpwic , Answer: 1.00 \\ [0.5ex] 
Step: 5, Question:  adnlong21 , Answer: 1.00 \\ [0.5ex] 
Step: 6, Ready to make a guess: Prob(y=1)=0.125, Guess: y=0, Ground truth: y=0 \\ [0.5ex] 
Episode terminated \\ [0.5ex] 

\noindent Starting new episode with a new test patient \\ [0.5ex] 
Basic info: sex: 2, age: 54, race:1 \\ [0.5ex] 
Step: 1, Question:  smkev1 , Answer: 0.00 \\ [0.5ex] 
Step: 2, Question:  phstat1 , Answer: 0.00 \\ [0.5ex] 
Step: 3, Question:  kidwkyr2 , Answer: 1.00 \\ [0.5ex] 
Step: 4, Question:  private2 , Answer: 0.00 \\ [0.5ex] 
Step: 5, Question:  livyr2 , Answer: 0.00 \\ [0.5ex] 
Step: 6, Question:  flwalk0 , Answer: 0.00 \\ [0.5ex] 
Step: 7, Ready to make a guess: Prob(y=1)=0.468, Guess: y=0, Ground truth: y=1 \\ [0.5ex] 
Episode terminated \\ [0.5ex] 

\noindent Starting new episode with a new test patient \\ [0.5ex] 
Basic info: sex: 1, age: 45, race:1 \\ [0.5ex] 
Step: 1, Question:  smkev1 , Answer: 0.00 \\ [0.5ex] 
Step: 2, Question:  phstat1 , Answer: 1.00 \\ [0.5ex] 
Step: 3, Question:  kidwkyr2 , Answer: 1.00 \\ [0.5ex] 
Step: 4, Question:  flsocl0 , Answer: 1.00 \\ [0.5ex] 
Step: 5, Question:  houseown2 , Answer: 0.00 \\ [0.5ex] 
Step: 6, Question:  flwalk0 , Answer: 1.00 \\ [0.5ex] 
Step: 7, Ready to make a guess: Prob(y=1)=0.076, Guess: y=0, Ground truth: y=0 \\ [0.5ex] 
Episode terminated \\ [0.5ex] 

\noindent Starting new episode with a new test patient \\ [0.5ex] 
Basic info: sex: 1, age: 77, race:0 \\ [0.5ex] 
Step: 1, Question:  smkev1 , Answer: 1.00 \\ [0.5ex] 
Step: 2, Question:  kidwkyr2 , Answer: 1.00 \\ [0.5ex] 
Step: 3, Question:  phstat1 , Answer: 0.00 \\ [0.5ex] 
Step: 4, Question:  vigfreqw , Answer: 95.00 \\ [0.5ex] 
Step: 5, Ready to make a guess: Prob(y=1)=0.906, Guess: y=1, Ground truth: y=1 \\ [0.5ex] 
Episode terminated \\ [0.5ex] 

\noindent Starting new episode with a new test patient \\ [0.5ex] 
Basic info: sex: 1, age: 80, race:1 \\ [0.5ex] 
Step: 1, Question:  smkev1 , Answer: 0.00 \\ [0.5ex] 
Step: 2, Question:  kidwkyr2 , Answer: 1.00 \\ [0.5ex] 
Step: 3, Question:  phstat1 , Answer: 0.00 \\ [0.5ex] 
Step: 4, Question:  phstat5 , Answer: 0.00 \\ [0.5ex] 
Step: 5, Question:  la1ar2 , Answer: 1.00 \\ [0.5ex] 
Step: 6, Question:  flwalk0 , Answer: 1.00 \\ [0.5ex] 
Step: 7, Ready to make a guess: Prob(y=1)=0.708, Guess: y=1, Ground truth: y=1 \\ [0.5ex] 
Episode terminated \\ [0.5ex] 

\noindent Starting new episode with a new test patient \\ [0.5ex] 
Basic info: sex: 1, age: 33, race:1 \\ [0.5ex] 
Step: 1, Question:  smkev1 , Answer: 0.00 \\ [0.5ex] 
Step: 2, Question:  phstat1 , Answer: 0.00 \\ [0.5ex] 
Step: 3, Question:  kidwkyr2 , Answer: 1.00 \\ [0.5ex] 
Step: 4, Question:  flwalk4 , Answer: 0.00 \\ [0.5ex] 
Step: 5, Question:  flwalk0 , Answer: 1.00 \\ [0.5ex] 
Step: 6, Ready to make a guess: Prob(y=1)=0.114, Guess: y=0, Ground truth: y=0 \\ [0.5ex] 
Episode terminated \\ [0.5ex] 

\noindent Starting new episode with a new test patient \\ [0.5ex] 
Basic info: sex: 2, age: 57, race:1 \\ [0.5ex] 
Step: 1, Question:  smkev1 , Answer: 1.00 \\ [0.5ex] 
Step: 2, Question:  phstat1 , Answer: 0.00 \\ [0.5ex] 
Step: 3, Question:  kidwkyr2 , Answer: 1.00 \\ [0.5ex] 
Step: 4, Question:  beddayr , Answer: 5.00 \\ [0.5ex] 
Step: 5, Question:  dibev1 , Answer: 0.00 \\ [0.5ex] 
Step: 6, Question:  amigr2 , Answer: 1.00 \\ [0.5ex] 
Step: 7, Ready to make a guess: Prob(y=1)=0.410, Guess: y=0, Ground truth: y=1 \\ [0.5ex] 
Episode terminated \\ [0.5ex] 

\noindent Starting new episode with a new test patient \\ [0.5ex] 
Basic info: sex: 1, age: 42, race:0 \\ [0.5ex] 
Step: 1, Question:  smkev1 , Answer: 0.00 \\ [0.5ex] 
Step: 2, Question:  phstat1 , Answer: 1.00 \\ [0.5ex] 
Step: 3, Question:  kidwkyr2 , Answer: 1.00 \\ [0.5ex] 
Step: 4, Question:  flsocl0 , Answer: 1.00 \\ [0.5ex] 
Step: 5, Question:  fliadlyn2 , Answer: 1.00 \\ [0.5ex] 
Step: 6, Question:  flwalk0 , Answer: 1.00 \\ [0.5ex] 
Step: 7, Ready to make a guess: Prob(y=1)=0.191, Guess: y=0, Ground truth: y=0 \\ [0.5ex] 
Episode terminated \\ [0.5ex] 

\noindent Starting new episode with a new test patient \\ [0.5ex] 
Basic info: sex: 2, age: 79, race:1 \\ [0.5ex] 
Step: 1, Question:  smkev1 , Answer: 0.00 \\ [0.5ex] 
Step: 2, Question:  kidwkyr2 , Answer: 1.00 \\ [0.5ex] 
Step: 3, Question:  phstat1 , Answer: 0.00 \\ [0.5ex] 
Step: 4, Question:  private2 , Answer: 1.00 \\ [0.5ex] 
Step: 5, Question:  la1ar2 , Answer: 1.00 \\ [0.5ex] 
Step: 6, Question:  livyr2 , Answer: 1.00 \\ [0.5ex] 
Step: 7, Ready to make a guess: Prob(y=1)=0.645, Guess: y=1, Ground truth: y=1 \\ [0.5ex] 
Episode terminated \\ [0.5ex] 

\noindent Starting new episode with a new test patient \\ [0.5ex] 
Basic info: sex: 2, age: 47, race:1 \\ [0.5ex] 
Step: 1, Question:  smkev1 , Answer: 1.00 \\ [0.5ex] 
Step: 2, Question:  phstat1 , Answer: 0.00 \\ [0.5ex] 
Step: 3, Question:  kidwkyr2 , Answer: 1.00 \\ [0.5ex] 
Step: 4, Question:  flsocl0 , Answer: 1.00 \\ [0.5ex] 
Step: 5, Question:  dibev1 , Answer: 0.00 \\ [0.5ex] 
Step: 6, Question:  amigr2 , Answer: 1.00 \\ [0.5ex] 
Step: 7, Ready to make a guess: Prob(y=1)=0.195, Guess: y=0, Ground truth: y=0 \\ [0.5ex] 
Episode terminated \\ [0.5ex] 

\noindent Starting new episode with a new test patient \\ [0.5ex] 
Basic info: sex: 1, age: 76, race:1 \\ [0.5ex] 
Step: 1, Question:  smkev1 , Answer: 1.00 \\ [0.5ex] 
Step: 2, Question:  kidwkyr2 , Answer: 1.00 \\ [0.5ex] 
Step: 3, Question:  phstat1 , Answer: 0.00 \\ [0.5ex] 
Step: 4, Question:  phstat5 , Answer: 0.00 \\ [0.5ex] 
Step: 5, Ready to make a guess: Prob(y=1)=0.839, Guess: y=1, Ground truth: y=1 \\ [0.5ex] 
Episode terminated \\ [0.5ex]

\end{document}